\documentclass[conference]{IEEEtran}
\IEEEoverridecommandlockouts

\usepackage{cite}
\usepackage{amsmath,amssymb,amsfonts}
\usepackage{algorithmic}
\usepackage{graphicx}
\usepackage{textcomp}
\usepackage{xcolor}
\def\BibTeX{{\rm B\kern-.05em{\sc i\kern-.025em b}\kern-.08em
    T\kern-.1667em\lower.7ex\hbox{E}\kern-.125emX}}
\usepackage{subcaption}
\usepackage{multirow}

\usepackage{listings}

\usepackage{pgfplots}
\usepackage{pgfplotstable}
\usepackage{xcolor}

\pgfplotsset{compat=1.18}

\begin{document}

\title{A Real-Time Tsetlin Machine-based Non-intrusive Load Monitoring System on MCUs
}


\author{
\IEEEauthorblockN{
Han Wu,
Tianhang Tan,
Shengyu Duan,
Alex Yakovlev,
Rishad Shafik and
Tousif Rahman
}
\IEEEauthorblockA{
\textit{School of Engineering}, \small{\textit{Newcastle University}, Newcastle upon Tyne, United Kingdom} \\
\small{Correspondence: han.wu2@newcastle.ac.uk}
}
}



\maketitle

\begin{abstract}
Non-Intrusive Load Monitoring (NILM) systems estimate individual appliance energy consumption from a single aggregate meter, without requiring separate sensors for each device. By installing a single meter that measures a building's total electricity consumption, NILM algorithms can determine the active status of each appliance. However, traditional NILM systems use computationally intensive optimization algorithms to process offline data, limiting their capability for on-device deployment, where sensitive household data must be processed locally. This paper proposes a Tsetlin Machine (TM)-based NILM framework, targeting real-time applications on resource-constrained microcontrollers (MCUs), enabling privacy-preserving edge deployment. The problem is reformulated as a classification task, and the proposed approach achieves an average precision of 90\% and recall of 96\% for two-appliance classification, and 77\% precision and 80\% recall for four appliances on the REDD dataset. The trained model occupies only 17 KB of flash memory and achieves an inference latency of 0.43 ms on an ESP32, demonstrating its suitability for NILM applications on MCUs.
\end{abstract}


\section{Introduction}
\begin{figure}[tb]
\centering

\begin{subfigure}{0.9\linewidth}
    \centering
    \includegraphics[width=\linewidth]{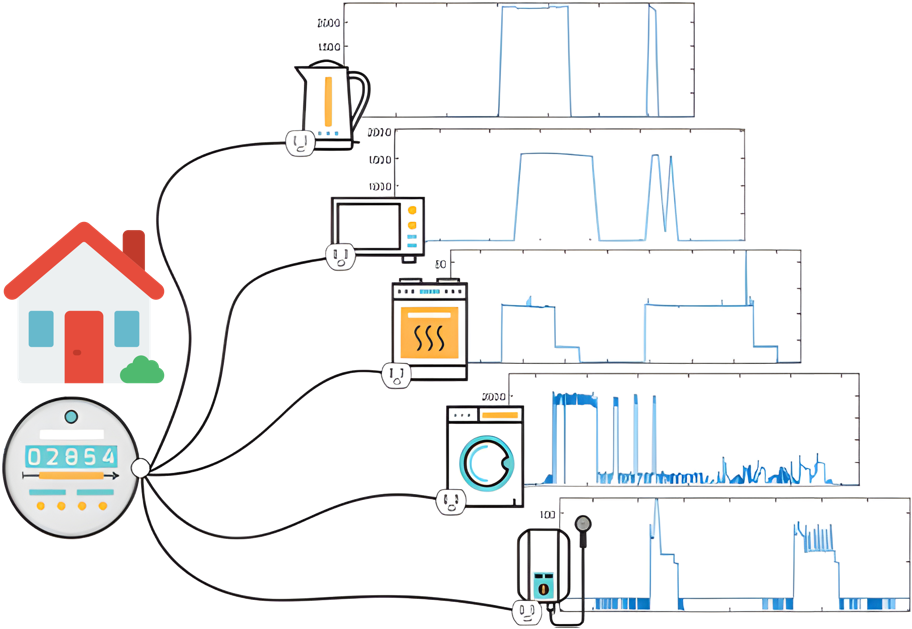}
    \caption{Illustration of an NILM system.}
    \label{fig:nilm}
\end{subfigure}

\vspace{0.5cm}

\begin{subfigure}{0.9\linewidth}
    \centering
    \includegraphics[width=\linewidth]{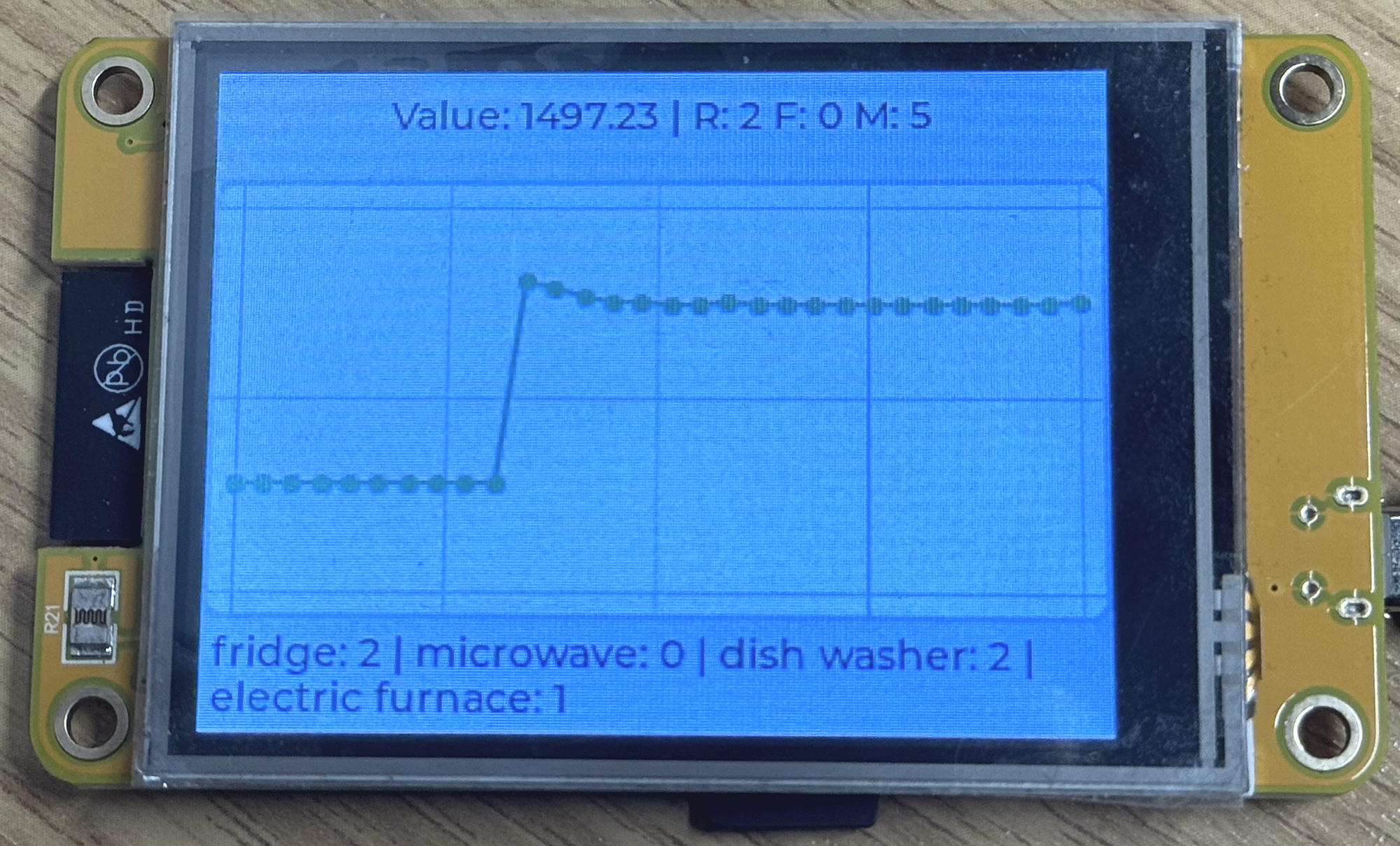}
    \caption{Demonstration of the proposed system on a MCU.}
    \label{fig:demo}
\end{subfigure}

\caption{(a) Non-Intrusive Load Monitoring (NILM) estimates the power consumption of individual appliances from a single aggregated mains meter without requiring dedicated sensors for each device. (b) Demonstration of the proposed TM-based NILM system on a MCU.}
\label{fig:nilm_demo}
\end{figure}

Non-intrusive Load Monitoring (NILM) aims to estimate the energy consumption of individual appliances within a building without installing a power meter for each device. Since deploying dedicated sensors for every appliance is both time-consuming and expensive, a more cost-effective approach is to install a single meter at the point where electricity enters the building (see Fig. \ref{fig:nilm}). Therefore, extracting the energy consumption of individual appliances from the total load becomes a key challenge, known as energy disaggregation.

\textbf{Traditional Model-based Methods}: The first NILM algorithm was proposed in \cite{hart1992nonintrusive}, where the energy disaggregation problem was formulated as a combinatorial optimization problem. In this approach, a Finite State Machine (FSM) is used to model the operating states of each appliance. Subsequently, a variety of model-based methods have been developed, including Factorial Hidden Markov Models (FHMMs) \cite{kolter2011redd} and Mixed-Integer Linear Programming (MILP) approaches \cite{wittmann2018nonintrusive}.

\textbf{Machine Learning-based Methods}: Model-based methods typically require prior knowledge of appliance characteristics and the number of appliances present in the household, which is often unavailable in real-world settings. To overcome these limitations, researchers turned to data-driven approaches that automatically extract appliance features from aggregated data. Both traditional Machine Learning (ML) models, such as k-Nearest Neighbor (KNN) and Support Vector Machine (SVM) \cite{tabanelli2021trimming}, as well as Deep Learning models \cite{piccialli2021improving, rajkumar2025non} are studied for the NILM problem.

\textbf{Real-time Applications}: Both Model-based methods and ML-based methods rely on computationally intensive optimization algorithms, which limit their applicability in real-time monitoring applications \cite{faustine2017survey}. 

To address this limitation, the main contributions of this work are as follows:

\begin{itemize}
    \item The NILM problem is reformulated as a classification task and addressed using the lightweight and memory-efficient Tsetlin Machine (TM) model \cite{granmo2018tsetlin}.
    \item A TM-based NILM system is developed to enable near real-time appliance-level energy disaggregation on resource-constrained MCUs.
    \item The implementation of the proposed TM-based NILM system is released as open source on GitHub\footnote{The source code is available at: https://github.com/wuhanstudio/nilm.}.
\end{itemize}


\begin{figure*}[tb]
\centering
\includegraphics[width=\linewidth]{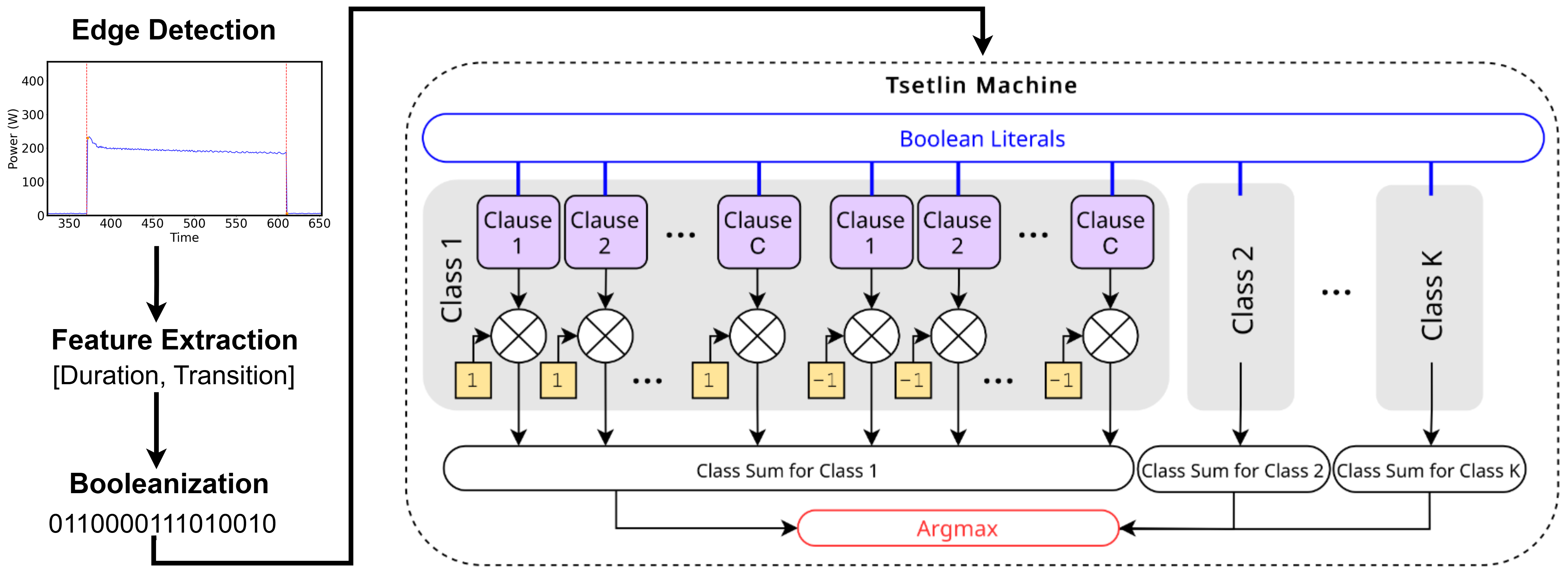}
\caption{The TM-based NILM approach extracts the 21 features from aggregated meter readings after edge detection and event pairing, normalizes and Booleanizes the resulting features, and then feeds them into the TM model for classification.}
\label{tm}
\end{figure*}

\section{TM-Based NILM System}

Due to the limited memory and computational power on MCUs, this work adopts the TM \cite{granmo2018tsetlin} as the underlying machine learning model. TM is a classification model that operates on binary input features, where each feature is a Boolean value (0 or 1) (see Fig. \ref{tm}). Without heavy floating-point multiplications, TM is more computationally efficient than neural networks and enables real-time NILM on MCUs. 

The TM-based NILM system consists of four main components: Edge Detection, Event Pairing, Feature Extraction, and TM-based Machine Learning classification:

\begin{itemize}
    \item \textbf{Edge Detection} module takes raw sensor data as input and identifies power transitions, which correspond to appliance activation (rising edges) and deactivation ( falling edges) events.
    \item \textbf{Event Pairing} algorithm groups corresponding rising and falling edges into complete appliance activity windows, which are then used for feature extraction.
    \item \textbf{Feature Extraction} module converts each appliance activity window into a compact feature representation, capturing transition characteristics, temporal patterns, steady-state behavior, and energy consumption information. 
    \item \textbf{TM-based Machine learning} module first converts extracted features to boolean values (either 0 or 1) and then classifies the appliance for each activity window.
\end{itemize}

\textbf{Edge Detection:} To reduce computational overhead on MCUs, the proposed NILM system adopts an event-driven architecture. Processing is triggered only when a transition is detected, that is, when the change in power consumption exceeds a predefined threshold, indicating that the power consumption has reached a new steady state. 

However, raw sensor data are inherently noisy, and transient spikes should not be detected as state transitions. To address this issue, Hart \cite{hart1992nonintrusive} proposed a steady-state detector that identifies only transitions between steady states, thereby filtering out transient noise. While the original implementation assumes access to offline data, the proposed system reimplements the detection algorithm for real-time streaming data on MCUs.

\textbf{Event Pairing:} After edge detection, each detected steady-state transition, whether a rising or falling edge, represents only a single change in power consumption. To capture a complete appliance activity, the corresponding rising and falling transitions must be paired to form an active-cycle event before feature extraction.  Features are then extracted from these paired event windows and used to train the TM model.

A local event-pairing algorithm keeps a list of recent rising edges. When a falling edge is detected, it searches this list and selects the rising edge, or a small group of rising edges, whose power change best matches the falling transition. This follows the common NILM assumption that the power increase when an appliance turns on should be close to the power decrease when it turns off. 

However, a local strategy can be sensitive to noisy or closely spaced transitions. Once a rising edge is paired or removed from the recent-edge list, the decision cannot be revised later. As a result, a nearby but incorrect rising edge may be selected, while a more suitable edge is skipped. 


To reduce this problem, the local search strategy is improved with a global event-pairing step. First, the rising and falling edges are sorted by time. A rising edge and a subsequent falling edge are considered as a candidate pair only when their duration is within a time limit. We then check whether their power changes are consistent. For a rising edge (i) and a falling edge (j), the power error is defined as
\begin{equation}
e_{ij}=|\Delta P_i^{+}+\Delta P_j^{-}|
\end{equation}

The pair is accepted only when this error is smaller than an allowed tolerance.

The tolerance combines an absolute and a relative margin:
\begin{equation}
\tau_{ij}=\max(100,;0.25|\Delta P_j^-|).
\end{equation}

This allows small measurement errors while still rejecting pairs with clearly inconsistent power changes. For each valid candidate pair, we use the score
\begin{equation}
s_{ij}=1-\frac{e_{ij}}{\tau_{ij}},
\end{equation}
where a higher score means a better ON/OFF power match. A dynamic program is then used to select the set of pairs with the highest total score. In this way, each detected edge is used at most once, and unreliable edges can remain unmatched.


\textbf{Feature Extraction:} To formulate NILM as a classification problem, features are extracted from each paired event, which defines an appliance activity window. For each detected appliance activation episode, a set of 21 features is extracted to characterize its transition behaviour, steady-state characteristics, temporal properties, and internal dynamics. The complete list of extracted features is summarized in Table~\ref{tab:episode_features}.





\begin{table}[t]
\centering
\caption{Features extracted from each paired appliance activity window.}
\label{tab:episode_features}
\begin{tabular}{ll}
\hline
\textbf{Category} & \textbf{Feature} \\
\hline
\multirow{8}{*}{Transition}
& Positive transition magnitude \\
& Negative transition magnitude \\
& Average absolute transition magnitude \\
& Logarithm of absolute transition magnitude \\
& Episode duration \\
& Logarithm of episode duration \\
& Transition--duration product \\
& Transition--duration ratio \\
\hline
\multirow{5}{*}{Steady-state statistics}
& Mean power \\
& Standard deviation of power \\
& Minimum power \\
& Maximum power \\
& Power range \\
\hline
\multirow{5}{*}{Internal dynamics}
& Mean absolute consecutive-sample difference \\
& Maximum absolute consecutive-sample difference \\
& Number of significant internal power transitions \\
& Estimated number of operating subcycles \\
& Active-state fraction \\
\hline
\multirow{3}{*}{Energy \& context}
& Episode energy estimate \\
& Post-event minus pre-event mean power \\
& Number of large internal power transitions \\
\hline
\end{tabular}
\end{table}

\textbf{TM-based Machine Learning:} A Tsetlin Machine (TM) is a rule-based machine learning model that learns patterns through collections of Tsetlin Automata (TAs). Each TA independently learns whether to include or exclude input literals, enabling the model to construct interpretable logical clauses. These clauses are combined to perform classification using simple Boolean operations, avoiding computationally expensive operations such as floating-point multiplication and matrix operations. This makes TM particularly suitable for deployment on MCUs.

\textbf{Booleanization}: TM accepts only Boolean values as input features. Therefore, each floating point feature is first normalized to the range $[0,1]$, then scaled to $[0,255]$ and converted to 8-bit integers. Lastly,  bitwise decomposition is applied to each integer, producing the Boolean literals for the TM.

\textbf{Model Inference}: The TM model is trained offline on a  computer and then deployed to the MCU for inference. The trained model is serialized using Protobuf, and the inference model retains only the indices of active Tsetlin Automata (TAs), eliminating unnecessary state information and substantially reducing the memory footprint.



All aforementioned components are implemented on the MCU, including edge detection, event pairing, feature extraction, and TM inference, ensuring that sensitive household data remains securely processed on local device.


\section{Experimental Results}


Experiments are conducted on the REDD dataset, which was collected from 6 residential buildings \cite{kolter2011redd} and pre-processed using the NILMTK toolkit \cite{batra2014nilmtk}. The dataset contains both aggregate measurements from the main meter and individual appliance-level power measurements. Buildings 1, 2, 4, 5, and 6 are used for training, while Building 3 is selected for testing because it contains measurements for all appliances considered in this study. Hyperparameter optimisation is performed using Optuna \cite{akiba2019optuna}. The TM is trained with 286 clauses, 196 TA states, and 8-bit feature encoding. The threshold ($T$) and specificity ($s$) parameters are set to 20 and 6.0, respectively. The model is trained for 10 epochs.

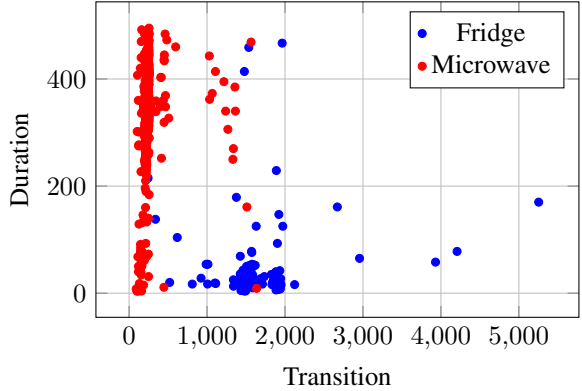
\begin{figure}[t]
\centering
\begin{tikzpicture}
\begin{axis}[
    width=0.9\linewidth,
    height=0.65\linewidth,
    xlabel={Transition},
    ylabel={Duration},
    grid=major,
    legend pos=north east,
]

\addplot[
    only marks,
    mark=*,
    mark size=1.5pt,
    blue,
]
table[
    x=transition,
    y=duration,
    col sep=comma,
    restrict expr to domain={\thisrow{label}}{1:1},
] {redd_scatter.csv};

\addlegendentry{Fridge}

\addplot[
    only marks,
    mark=*,
    mark size=1.5pt,
    red,
]
table[
    x=transition,
    y=duration,
    col sep=comma,
    restrict expr to domain={\thisrow{label}}{0:0},
] {redd_scatter.csv};

\addlegendentry{Microwave}

\end{axis}
\end{tikzpicture}
\caption{Scatter plot of appliance events in the Transition-Duration feature space.}
\label{fridge}
\end{figure}

\textbf{Binary Classification:} First, a binary classification TM model was trained to distinguish between the active cycle of a fridge and a microwave. As shown in Tab.~\ref{tm_binary}, the model achieved high precision and recall for both appliances. The two appliances are easily separable because the fridge cycles are characterized by low-power transitions and long operating durations, whereas microwave cycles typically involve high-power transitions over short durations (see Fig.\ref{fridge}).

\begin{table}[tb]
\centering
\caption{Precision and Recall of the TM trained on Fridge and Microwave}
\begin{tabular}{lcc}
\hline
\textbf{Appliance} & \textbf{Precision} & \textbf{Recall} \\
\hline
Fridge     & 0.99 & 0.97 \\
Microwave  & 0.80 & 0.95 \\
\hline
\end{tabular}
\label{tm_binary}
\end{table}

\begin{table}[tb]
\centering
\caption{Precision and Recall of the TM trained on multiple appliances}
\begin{tabular}{lcc}
\hline
\textbf{Appliance} & \textbf{Precision} & \textbf{Recall} \\
\hline
Fridge            & 0.84 & 0.94 \\
Microwave         & 0.75 & 0.95 \\
Dish Washer       & 0.10 & 0.08 \\
Electric Furnace  & 0.50 & 0.15 \\
\hline
\end{tabular}
\label{tm_multi}
\end{table}

\begin{table}[tb]
\centering
\caption{F1 Score and Model Size of Models Trained on REDD Dataset}
\begin{tabular}{lccc}
\hline
\textbf{Method} & \textbf{Fridge} & \textbf{Microwave} & \textbf{Model Size} \\
\hline
Our Method & 0.89 & 0.84 & 17 KB \\
LSTM \cite{mauch2015new} & 0.92 & 0.38 & 1.9 MB \\
BERT \cite{yue2020bert4nilm} & 0.76 & 0.48 & 8.4 MB \\
EdgeNILM \cite{kukunuri2020edgenilm} & 0.74 & - & 41.5 MB \\
\hline
\end{tabular}
\label{tm_sota}
\end{table}


\textbf{Multi-class Classification:} However, as the number of appliances increased, the overall classification performance decreased due to more complex operating patterns of certain appliances, as shown in Tab.~\ref{tm_multi}. For example, the electric furnace often consist of multiple short active sub-cycles within a single operating cycle, making it difficult to distinguish from other appliances.

\textbf{SRAM and Flash Memory Consumption:} As shown in Table~\ref{tm_sota}, the proposed TM model achieves competitive F1 scores while requiring substantially less memory than the compared models. The trained TM model requires only 17~KB of Flash memory and 168~bytes of SRAM, corresponding to 21 features encoded with 8 bits each. It achieves an inference time of approximately 0.43~ms per sample on an ESP32 using the Arduino framework, demonstrating its suitability for real-time deployment on MCUs.

\section{Discussion}

The proposed TM-based NILM approach has several limitations. The main bottleneck lies in the accuracy of edge detection and event pairing, as incorrect detection of power transitions or failure to correctly associate rising and falling edges can lead to inaccurate appliance activity windows and consequently degrade classification performance. In addition, the performance decreases in multi-class scenarios due to the overlapping and complex power signatures of different appliances. Appliances with similar operating patterns or multi-stage operating cycles are particularly challenging to distinguish using the extracted Transition-Duration features. Future work will investigate more robust event detection and pairing strategies, as well as richer feature representations, to improve scalability to a larger number of appliances.


\section{Conclusion}
In conclusion, this paper presents a TM-based approach for NILM on resource-constrained MCUs. By reformulating NILM as a classification problem, the proposed method achieves an average precision of 90\% and recall of 96\% for two-appliance classification on the REDD dataset, while attaining 77\% precision and 80\% recall for four appliances. Furthermore, the trained model requires only 17 KB of Flash and achieves an inference speed of 0.43 ms per sample on an ESP32, making it suitable for real-time embedded applications.

\section{Acknowledgment}
This research was funded through the EPSRC Northern Net Zero Accelerator  (NNZA) EP/Y024052/1 for the project “NNZA: ECO-HAVEN: Energy Conservation for Households using Adaptive Edge Learning”. 

\bibliographystyle{IEEEtran}
\bibliography{IEEEabrv, mybibfile}

\begin{thebibliography}{10}
\providecommand{\url}[1]{#1}
\csname url@samestyle\endcsname
\providecommand{\newblock}{\relax}
\providecommand{\bibinfo}[2]{#2}
\providecommand{\BIBentrySTDinterwordspacing}{\spaceskip=0pt\relax}
\providecommand{\BIBentryALTinterwordstretchfactor}{4}
\providecommand{\BIBentryALTinterwordspacing}{\spaceskip=\fontdimen2\font plus
\BIBentryALTinterwordstretchfactor\fontdimen3\font minus \fontdimen4\font\relax}
\providecommand{\BIBforeignlanguage}[2]{{%
\expandafter\ifx\csname l@#1\endcsname\relax
\typeout{** WARNING: IEEEtran.bst: No hyphenation pattern has been}%
\typeout{** loaded for the language `#1'. Using the pattern for}%
\typeout{** the default language instead.}%
\else
\language=\csname l@#1\endcsname
\fi
#2}}
\providecommand{\BIBdecl}{\relax}
\BIBdecl

\bibitem{hart1992nonintrusive}
G.~W. Hart, ``Nonintrusive appliance load monitoring,'' \emph{Proceedings of the IEEE}, vol.~80, no.~12, pp. 1870--1891, 1992.

\bibitem{kolter2011redd}
J.~Z. Kolter and M.~J. Johnson, ``Redd: A public data set for energy disaggregation research,'' in \emph{Workshop on data mining applications in sustainability (SIGKDD), San Diego, CA}, vol.~25, 2011, pp. 59--62.

\bibitem{wittmann2018nonintrusive}
F.~M. Wittmann, J.~C. L{\'o}pez, and M.~J. Rider, ``Nonintrusive load monitoring algorithm using mixed-integer linear programming,'' \emph{IEEE Transactions on Consumer Electronics}, vol.~64, no.~2, pp. 180--187, 2018.

\bibitem{tabanelli2021trimming}
E.~Tabanelli, D.~Brunelli, A.~Acquaviva, and L.~Benini, ``Trimming feature extraction and inference for mcu-based edge nilm: A systematic approach,'' \emph{IEEE Transactions on Industrial Informatics}, vol.~18, no.~2, pp. 943--952, 2021.

\bibitem{piccialli2021improving}
V.~Piccialli and A.~Sudoso, ``Improving non-intrusive load disaggregation through an attention-based deep neural network. energies 2021, 14, 847,'' \emph{Energy Data Analytics for Smart Meter Data}, p.~63, 2021.

\bibitem{rajkumar2025non}
N.~Rajkumar, A.~Govindaram, R.~Geetha, S.~ASF \emph{et~al.}, ``Non-intrusive load monitoring techniques for intelligent energy management: A comparative study of fhmm and lstm approach,'' in \emph{2025 International Conference on Visual Analytics and Data Visualization (ICVADV)}.\hskip 1em plus 0.5em minus 0.4em\relax IEEE, 2025, pp. 869--874.

\bibitem{faustine2017survey}
A.~Faustine, N.~H. Mvungi, S.~Kaijage, and K.~Michael, ``A survey on non-intrusive load monitoring methodies and techniques for energy disaggregation problem,'' \emph{arXiv preprint arXiv:1703.00785}, 2017.

\bibitem{granmo2018tsetlin}
O.-C. Granmo, ``The tsetlin machine--a game theoretic bandit driven approach to optimal pattern recognition with propositional logic,'' \emph{arXiv preprint arXiv:1804.01508}, 2018.

\bibitem{batra2014nilmtk}
N.~Batra, J.~Kelly, O.~Parson, H.~Dutta, W.~Knottenbelt, A.~Rogers, A.~Singh, and M.~Srivastava, ``Nilmtk: An open source toolkit for non-intrusive load monitoring,'' in \emph{Proceedings of the 5th international conference on Future energy systems}, 2014, pp. 265--276.

\bibitem{akiba2019optuna}
T.~Akiba, S.~Sano, T.~Yanase, T.~Ohta, and M.~Koyama, ``{O}ptuna: A next-generation hyperparameter optimization framework,'' in \emph{The 25th ACM SIGKDD International Conference on Knowledge Discovery \& Data Mining}, 2019, pp. 2623--2631.

\bibitem{mauch2015new}
L.~Mauch and B.~Yang, ``A new approach for supervised power disaggregation by using a deep recurrent lstm network,'' in \emph{2015 IEEE global conference on signal and information processing (GlobalSIP)}.\hskip 1em plus 0.5em minus 0.4em\relax IEEE, 2015, pp. 63--67.

\bibitem{yue2020bert4nilm}
Z.~Yue, C.~R. Witzig, D.~Jorde, and H.-A. Jacobsen, ``Bert4nilm: A bidirectional transformer model for non-intrusive load monitoring,'' in \emph{Proceedings of the 5th international workshop on non-intrusive load monitoring}, 2020, pp. 89--93.

\bibitem{kukunuri2020edgenilm}
R.~Kukunuri, A.~Aglawe, J.~Chauhan, K.~Bhagtani, R.~Patil, S.~Walia, and N.~Batra, ``Edgenilm: Towards nilm on edge devices,'' in \emph{Proceedings of the 7th ACM international conference on systems for energy-efficient buildings, cities, and transportation}, 2020, pp. 90--99.

\end{thebibliography}

\end{document}